\documentclass{article} 
\usepackage{iclr2024_conference,times}

\usepackage{amsmath,amsfonts,bm}

\def\eqref#1{equation~\ref{#1}}

\def\1{\bm{1}}

\DeclareMathAlphabet{\mathsfit}{\encodingdefault}{\sfdefault}{m}{sl}
\SetMathAlphabet{\mathsfit}{bold}{\encodingdefault}{\sfdefault}{bx}{n}

\usepackage{wrapfig}
\usepackage{hyperref}
\usepackage{url}
\usepackage{epsfig}
\usepackage{arydshln} 
\usepackage{color}

\usepackage{algorithm}
\usepackage{algorithmic}
\usepackage{multirow}

\title{\textbf{TEAL}: \textbf{T}okenize and \textbf{E}mbed \textbf{AL}l for multi-modal large language models}

\author{
\centerline{Zhen Yang\footnotemark[1] ~~ Yingxue Zhang\footnotemark[1] ~~ Fandong Meng ~~ Jie Zhou} \\
 \centerline{\normalfont{WeChat AI, Tencent Inc.}} \\
 \centerline{\texttt{\{zieenyang,yxuezhang,fandongmeng,withtomzhou\}@tencent.com}}
}

\iclrfinalcopy 
\begin{document}

\footnotetext[1]{These authors contribute equally to this work.}
\maketitle

\begin{abstract}
Despite Multi-modal Large Language Models (MM-LLMs) have made exciting strides recently, they are still struggling to efficiently model the interactions among multi-modal inputs and the generation in non-textual modalities. In this work, we propose \textit{TEAL (Tokenize and Embed ALl)}, an approach to treat the input from any modality as a token sequence and learn a joint embedding space for all modalities. Specifically, for the input from any modality, \textit{TEAL} firstly discretizes it into a token sequence with the off-the-shelf tokenizer and embeds the token sequence into a joint embedding space with a learnable embedding matrix. MM-LLMs just need to predict the multi-modal tokens autoregressively as conventional textual LLMs do.  Finally, the corresponding de-tokenizer is applied to generate the output in each modality based on the predicted token sequence. With the joint embedding space, \textit{TEAL} enables the frozen LLMs to perform both understanding and generation tasks involving non-textual modalities, such as image and audio. Thus, the textual LLM can just work as an interface and maintain its high performance in textual understanding and generation. Experiments show that \textit{TEAL} achieves substantial improvements in multi-modal understanding, and implements a simple scheme for multi-modal generations.

\end{abstract}

\section{Introduction}
Recently, Multi-Modal Large Language Models (MM-LLMs), which perform understanding and generation tasks more than textual modalities, have made exciting strides and garnered significant attention for their potential in Artificial Intelligence Generated Content (AIGC) \citep{cao2023comprehensive}.  MM-LLMs are considered a step closer to Artificial General Intelligence (AGI) \citep{goertzel2007artificial,fei2022towards} due to their provision of more user-friendly interfaces and their ability to perceive the world similarly to humans \citep{yin2023survey}. Typically, there are two main different branches in the realm of constructing MM-LLMs: One branch aims to construct a `real` multi-modal model by training the model with multi-modal data from scratch, without relying on the pre-trained textual LLMs \citep{borsos2023audiolm,lu2022unified,barrault2023seamlessm4t,shukor2023unified,chen2023pali3,copet2023simple}; The other branch takes the textual LLMs as the backbone and enables them to perform multi-modal understanding and generation tasks with instruction tuning. 

With the rapid advancement of textual LLMs, researchers are keener on the second branch of approaches which empowers the pre-trained high-performance textual LLMs with multi-modal abilities. In this line, some typical works, such as BLIP-2 \citep{li2023blip}, Flamingo \citep{alayrac2022flamingo}, MiniGPT-4 \citep{zhu2023minigpt}, LLama-Adapter \citep{gao2023llama,zhang2023llama}, LLaVA \citep{liu2023visual,liu2023improved}, SpeechGPT \citep{zhang2023speechgpt}, involve employing adapters that align pre-trained encoders in other modalities to textual LLMs. As these works take the dense features from the pre-trained encoders as additional non-textual information, they cannot efficiently model the interactions among multi-modal inputs and falter in the nuanced art of generating non-textual content. In order to compensate for this deficiency in the non-textual generation, some efforts, such as visual-ChatGPT \citep{chen2023pali3}, Hugging-GPT \citep{shen2023hugginggpt}, Audio-GPT \citep{huang2023audiogpt}, Next-GPT \citep{wu2023next}, and MiniGPT-5 \citep{zheng2023minigpt} have sought to amalgamate the textual LLMs with some external generation tools, e.g., Stable Diffusion \citep{rombach2022high}, DALL-E \citep{ramesh2021zeroshot}, Whisper \citep{radford2023robust}. Unfortunately, these systems suffer from two critical challenges due to their complete pipeline architectures. First, the information transfer between different modules is entirely based on generated textual tokens, where the process may lose some multi-modal information and propagate errors \citep{wu2023next}. Additionally, the external tools usually make the models complex and heavy, which consequently results in inefficient training and inference.

Based on the above observation, we conclude that the emerging challenges in the previous works are mainly raised by their non-unified processing of the multi-modal inputs, where they encode the non-textual inputs into a dense and high-level feature, but tokenize the textual input into a token sequence. The non-unified processing introduces an extra burden for LLMs to model the interaction between multi-modal inputs and generate the non-textual samples. In a nutshell, if we can tokenize the interleaved multi-modal input into a token sequence and align the non-textual token embedding into the textual embedding space, the original textual LLMs can be easily transformed to handle non-textual understanding and generation tasks with parameters tuned as little as possible.

In pursuit of this goal and inspired by the recent advancement of multi-modal tokenizers \citep{yu2023language,chang2023exploring,peng2022beit,borsos2023audiolm,yu2023spae}, we propose \textit{TEAL}, a token-in-token-out MM-LLM designed to seamlessly handle the token input and output in any combination of three modalities: text, image, and audio. Specifically, \textit{TEAL} comprises three tiers. First, we tokenize the input from any modality into a token sequence with the off-the-shelf tokenizers, such as BEiT-V2 and a Whisper-based audio tokenizer.  Second, we insert a non-textual embedding matrix and output matrix into an open-source textual LLM, which enables the textual LLM to process the non-textual inputs and outputs. To align the non-textual embedding matrices with their textual counterparts, we equip them with a projection layer. Third, the generated tokens are routed to the corresponding de-tokenizers, which transform the token sequences into samples in different modalities. We conduct extensive experiments on the modalities of text, image, and audio. Experimental results show that \textit{TEAL} achieves substantial improvements over previous works on multi-modal understanding and paves a simple way for the generation of non-textual modalities.



In summary, our contributions are three-fold:
\begin{enumerate}
\item{We propose \textit{TEAL}, an approach that treats the input from any modality as a token sequence and learns a joint embedding space for all modalities. \textit{TEAL} introduces a simple way to enable the frozen LLMs to perform both understanding and generation tasks involving non-textual modalities.}
\item{We conduct extensive experiments on the non-textual modalities of image and audio. Experimental results show that \textit{TEAL} achieves substantial improvements over previous works on multi-modal understanding and paves a simple way for the generation of non-textual modalities. To the best of our knowledge, this is the first work that successfully empowers the frozen LLM to perform tasks involving both the non-textual modalities of audio and image. }
\item{By testing versatile tokenizers for image and audio, we find that the tokenizer is key to the performance of MM-LLMs. Our extensive experiments have identified a new research direction that devising a general semantic-aware tokenizer is very promising.}
\end{enumerate}

\section{Related Work}
\subsection{MM-LLMs}
Training a multi-modal large language model from scratch in an end-to-end manner incurs substantial costs. Therefore, most researchers choose to integrate multi-modal modules into existing text-based large language models, allowing these models to acquire multi-modal capabilities. One branch involves employing robust pre-trained vision or audio encoders to encode multi-modal information into features and subsequently align it with the feature space of an LLM \citep{dai2023instructblip, chen2023xllm, zhang2023videollama, zhang2023llama, gao2023llama, ling2023adapting, wu2023decoderonly, hussain2023m}.  For example, Flamingo \citep{alayrac2022flamingo} utilizes vision encoders to obtain a fixed number of visual tokens and use cross-attention
layers to connect the pre-trained LLM layers. BLIP-2 \citep {li2023blip} utilizes a Q-Former as a bridge between the input image and the LLMs. LauraGPT \citep{chen2023lauragpt} uses a pre-trained Conformer-based encoder to extract continuous audio representations for the connected LLM. Furthermore, different projection layers are used to reduce the modality gap, such as a simple Linear Layer \citep{liu2023improved} or a two-layer Multi-layer Perceptron \citep{zhang2023pmcvqa}.
Moreover, LLaMa-Adapter \citep{zhang2023llama,gao2023llama} integrates trainable adapter modules into LLMs, enabling effective parameter tuning for the fusion of multi-modal information. Another branch involves using off-the-shelf expert models to convert images or speech into natural language in an offline manner, such as Next-GPT \citep{wu2023next}, SpeechGPT \citep{zhang2023speechgpt} and AudioGPT \citep{huang2023audiogpt}. 

Contrary to these works mentioned above, we tokenize the input from any modality into a token sequence and train a token-in-token-out MM-LLM designed to seamlessly handle the token input and output in any combination of three modalities: text, image, and audio.

\subsection{Non-textual discretization}
In addition to directly integrating multi-modal modules or using offline expert models, there are also efforts focused on non-textual discretization, which employs tokenizers to convert continuous 
images or audio into token sequences. This way, all modalities share the same form as tokens, which can be better compatible with LLM. 
Next, we will introduce two mainstream methods of Non-textual discretization.

\paragraph{VQ-VAEs}
Vector Quantised Variational AutoEncoder (VQ-VAE) \citep{van2017neural} is a seminal contribution in the field of non-textual tokenization, which incorporates vector quantization (VQ) to learn discrete representations and converts images into a sequence of discrete codes. In the vision domain, VQGAN \citep{esser2021taming} follows the idea, using a codebook to discretely encode images, and employs Transformer as the encoder. ViT-VQGAN \citep{yu2021vector} introduces several enhancements to the vanilla VQGAN, encompassing architectural modifications and advancements in codebook learning. BEiT-V2 \citep{peng2022beit} proposes Vector-quantized Knowledge Distillation (VQ-KD) to train a semantic-rich visual tokenizer by reconstructing high-level features from the teacher model. \citet{ge2023planting} proposes SEED and claims two principles for the tokenizer architecture and training that can ease the alignment with LLMs. \citet{yu2023spae} introduce SPAE, which can convert between raw pixels and lexical tokens extracted from the LLM’s vocabulary, enabling frozen LLMs to understand and generate images or videos.  For the audio, \citet{dieleman2018challenge} utilize autoregressive discrete autoencoders (ADAs) to capture correlations in waveforms. Jukebox \citep{dhariwal2020jukebox} uses a multi-scale VQ-VAE to compress music to discrete codes and model those using autoregressive Transformers, which can generate music with singing in the raw audio domain. SoundStream \citep{Zeghidour2021SoundStreamAE} employs a model architecture composed of a fully convolutional encoder/decoder network and adopts a Residual Vector Quantizer (RVQ) to project the audio embedding in a codebook of a given size. \citet{defossez2022high}, \citet{jiang2022crossscale} also adopt RVQ to quantize the output of the encoder. 
\paragraph{Clustering} 
Except for those methods that use trained specialized vector quantization (VQ) modules as tokenizers, some works \citep{lakhotia-etal-2021-generative,kharitonov-etal-2022-text} apply the clustering algorithms to the features, and the cluster indices are directly used as the discrete tokens for speech. The cluster approach typically relies on self-supervised learning models, such as HuBERT \citep{hsu2021hubert}, W2V-BERT \citep{chung2021w2v, borsos2023audiolm}, USM \citep{zhang2023google, rubenstein2023audiopalm}, which are trained for discrimination or masking prediction and maintain semantic information of the
speech. Compared with neural VQ-based tokenizers, the clustering-based approach provides enhanced flexibility as it can be applied to any pre-trained speech model without altering its underlying model structure.

\section{Method}
\begin{figure}
    \centering
    \resizebox{1.0\textwidth}{!}{
    \includegraphics{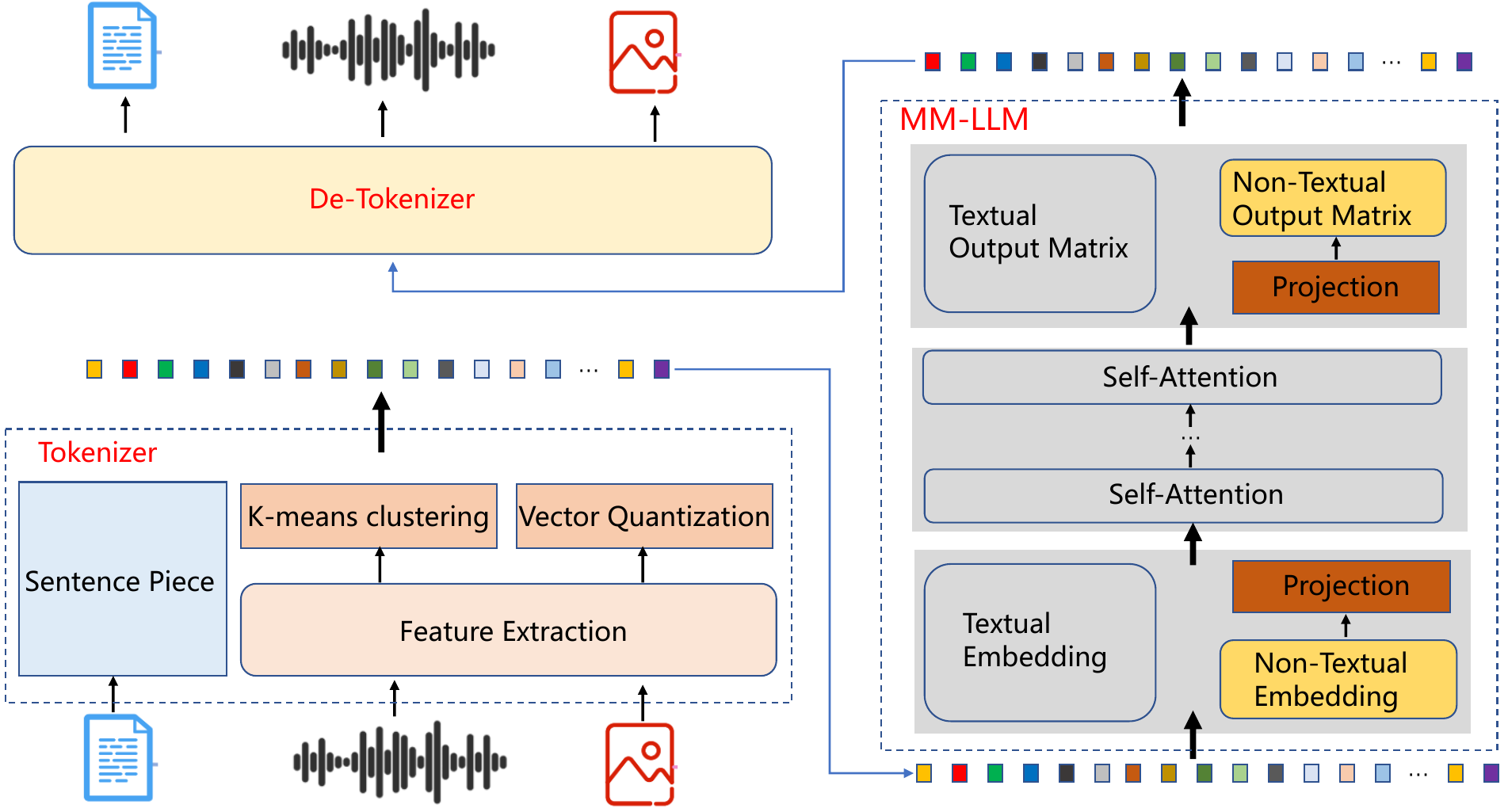}}
    \caption{The main architecture of \textit{TEAL}. The modules in MM-LLM denoted with the color gray make up the original textual LLM and most of them are frozen during training.}
    \label{Model_Arch}
\end{figure}

The main goal of this paper is to enable the frozen textual LLMs to model sequences consisting of multi-modal discrete tokens. Thus, the textual LLMs obtain the ability to perform both understanding and generation tasks involving non-textual modalities and maintain their strong abilities in text. The main architecture of our method is illustrated in Figure \ref{Model_Arch}. Firstly, we discretize the interleaved multi-modal input into a token sequence with the off-the-shelf tokenizers. Then, an open-sourced textual LLM is used to model the input and output token sequence by aligning the textual and non-textual embedding space. Finally, the corresponding off-the-shelf decoder is utilized to generate the output in each modality. In the remainder of this section, we will describe the model architecture in Subsection \ref{model_arch}. The tokenizer and de-tokenizer for non-textual modalities we used in this paper will be presented in Subsection \ref{tokenizer}. Finally, we propose our two-stage training strategies in Subsection \ref{training_strategy}. 

\subsection{Model Architecture}
\label{model_arch}
\textit{TEAL} is a general method that can be applied to any open-source LLMs. In this paper, the proposed MM-LLM takes the most popular open-sourced textual LLM, i.e., LLaMA, as the backbone, which makes it easy to compare fairly with previous works. To support the modeling of non-textual tokens, the MM-LLM also incorporates a non-textual embedding layer and a non-textual output layer. Two projection layers are applied after the non-textual embedding layer and before the output layer separately, which mainly serve two purposes: 1)  make the output dimension of textual and non-textual embedding the same; 2) align the non-textual embedding with the textual embedding space. To ease the training process and solve the cold-start problem, we initialize the non-textual embedding and output matrix with the codebook of the tokenizer, which will be described in Subsection \ref{tokenizer} in detail.

\subsection{Tokenize and De-Tokenize}
\label{tokenizer}
Tokenization is a very popular technique in the area of natural language processing, which is usually used as a tool to split the input sentence into the granularity of sub-words. Most of the existing textual LLMs take the sentence piece as the tokenizer for its universal processing of multi-lingual texts. The de-tokenization for the sentence piece is very simple, which just works as a function to replace the meta-symbol `$\_$' with the whitespace. Recently, tokenization (or denoted as discretization) in non-textual modalities has gained much attention and achieved substantial improvements, which makes it possible to build a fully token-in-token-out MM-LLM. The most widely used methods are VQ-VAE and k-means clustering. In this paper, we take the encoder of the VQ-VAE models and the k-means clustering as the tokenizers for the image and audio respectively. The decoders of the VQ-VAE models are taken as the de-tokenizers for the image and audio. For the image, we test the following typical tokenizers (and the corresponding de-tokenizers):
\begin{itemize}
    \item DALL-E \citep{ramesh2021zeroshot}: They train a discrete variational autoen-coder (dVAE) to compress each 256×256 RGB image into a 32 × 32 grid of image tokens, each element of which can assume 8192 possible values. We harness the open-source toolkit implemented by DALLE-pytorch.\footnote{https://github.com/lucidrains/DALLE-pytorch}. 
    \item VQ-GAN \citep{esser2021taming}: They combine the efficiency of convolutional approaches with the expressivity of transformers by introducing a convolutional VQGAN, which learns a codebook of context-rich visual parts, whose composition is modeled with an autoregressive transformer. We follow the open-source toolkit, Taming-Transformer, and directly use their released pre-trained models.\footnote{https://github.com/CompVis/taming-transformers}
    \item BEiT-V2 \citep{peng2022beit}: They propose vector-quantized knowledge distillation (VQ-KD) to train the visual tokenizer, where the tokenizer is trained to reconstruct the semantic features of a teacher model. We utilize the officially released toolkit and models.\footnote{https://github.com/microsoft/unilm}
\end{itemize}
For the audio, we apply K-means Clustering on the intermediate features of the following typical models, and the cluster indices are directly used as the discrete tokens for speech.
\begin{itemize}
    \item HuBERT \citep{hsu2021hubert}:  They incorporate an offline clustering step to generate aligned target labels for a BERT-like prediction loss for self-supervised representation learning. Through masked prediction, the model is forced to learn both acoustic and language models from continuous inputs.
    \item Whisper \citep{radford2023robust}: Whisper is a Transformer-based speech recognition model, which is trained on many different speech processing tasks via large-scale weak multilingual and multitask supervision. In this paper, we conduct experiments with the $Whisper_{small}$ to get discrete audio tokens.
    
\end{itemize}

\subsection{Two-stage Supervised Finetuning}
\label{training_strategy}
The proposed \textit{TEAL} model is initialized with the open-sourced textual LLM. To obtain the understanding and generation ability in non-textual modalities and maintain its high performance in textual modality, we propose a two-stage supervised fine-tuning that trains the model with parameters tuned as little as possible. In the following, we denote the two stages of supervised fine-tuning as pre-training and fine-tuning separately.
\paragraph{Pre-training} The goal of the pre-training is to align the non-textual and textual embedding space by tuning the projection layer. Specifically, we freeze all parameters in the MM-LLM except the parameter of the two projection layers. We generate the training samples from the vision-language and audio-language pairs with very simple prompts. Taking the vision-language pair as an example, we generate two training samples from each vision-language pair with the following format:
$$\text{The image and text pair:} [\text{img}] [\text{text}]$$
$$\text{The text and image pair:} [\text{text}]  [\text{img}]$$

\paragraph{Fine-tuning} In the stage of fine-tuning, we process the corpus of downstream tasks as the prompt format in \cite{zhang2023llama}. For each task, we use the GPT4 to generate 10 different prompts.\footnote{For details of the prompt format, we refer the readers to the Appendix \ref{sec_prompt}.} We freeze the parameters of the textual LLM and tune all parameters related to the non-textual modalities. Following \cite{zhang2023llama}, we apply the bias-norm tuning where the bias and norm parameters are inserted in each layer to enhance the fine-tuning performance. We also tested Lora tuning, but we did not obtain further improvement.

\section{Experiments}
We first test our method on the understanding tasks involving non-textual modalities, i.e., the task of coco-caption, science-QA, and CoVoST 2. Then, we report our performance on the task of image generation. The model is implemented based on the codebase of LLaMA-Adapter \citep{gao2023llama}.\footnote{https://github.com/Alpha-VLLM/LLaMA2-Accessory} If there is no specific explanation, all models are trained with two-stage supervised fine-tuning with 8 A100 GPUs, and the main hyper-parameters are set the same with LlaMA-Adapter. Following \citep{gao2023llama}, we also adopt top-p sampling as the default decoding method with
a temperature of 0.1 and a top-p of 0.75.

\subsection{COCO-Caption}
\begin{figure}[t]
    \centering
    \resizebox{1.0\textwidth}{!}{
    \includegraphics{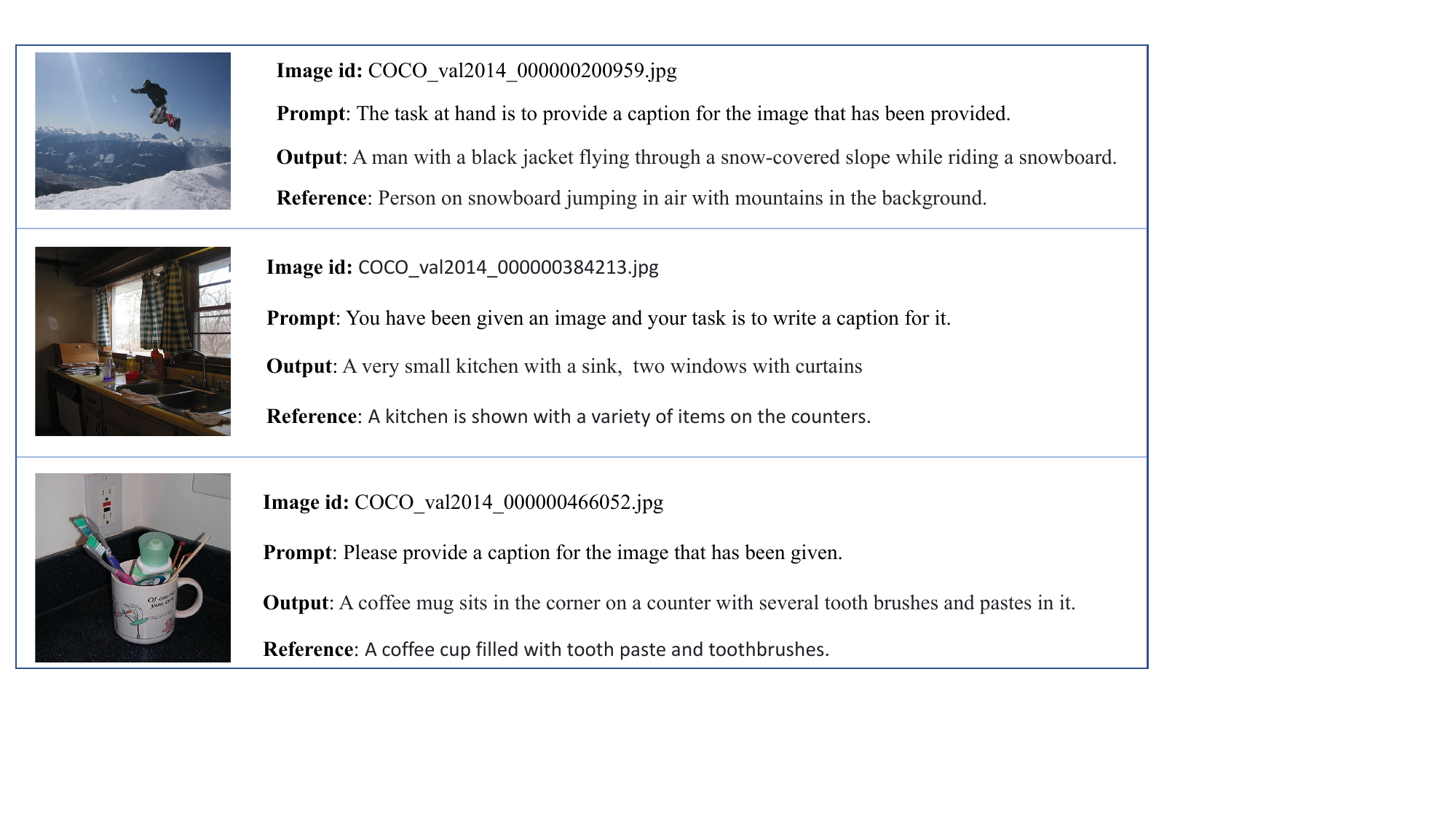}}
    \caption{Some examples in the coco2014 validation set. For each case, we present the original image ID, the prompt, the output of our model, and one reference caption randomly selected among all five references.}
    \label{caption_case}
\end{figure}

We utilize all image-caption pairs from the coco2014 dataset \citep{chen2015microsoft}, which contains 83K images for training. As there are at least five captions for each image in the coco2014 dataset, we can construct at least five training examples for each image by pairing the image with its all captions respectively. For a fair comparison, we report the CIDER, BLEU-4 on the Karpathy test split, which is evaluated with the official toolkit, pycocoeval.\footnote{https://github.com/cocodataset/cocoapi} The result is presented in Table \ref{Caption}. From Table \ref{Caption}, we can find that the proposed \textit{TEAL} achieves substantial improvements compared to the baseline of LLaMA-Adapter v2, which applies a frozen vision encoder to incorporate the vision information. Specifically, we achieve 1.9 and 6.6 points improvement on the metrics of BLEU-4 and CiDER respectively. Additionally, compared to the models that trained with large-scale corpora, such as the BLIP and BLIP2, \textit{TEAL} further narrows the performance gap without additional pre-training corpus. The cases on the valid set are shown in Figure \ref{caption_case}. We can find that the proposed \textit{TEAL} is able to understand the content of images well and can describe the details of the images clearly.

\begin{table}
\centering
\begin{tabular}{l|cc|cc}
\hline
   \multicolumn{1}{c}{\multirow{2}{*}{Model}} & \multicolumn{2}{c}{Data Scale} & \multicolumn{2}{c}{COCO Caption} \\
   \multicolumn{1}{c|}{} & PT & FT & CiDER&	BLEU-4       \\
    \hline
 LlaMA-Adapter v2 \citep{gao2023llama} & 0 & 0.6M & 122.2&  36.2  \\
 \hdashline
BLIP \citep{li2022blip} & 14M & 0.6M& 136.7 & 40.4 \\
BLIP2 \citep{li2023blip} & 129M& 0.6M& 145.3 & 43.7 \\
\hdashline
\textbf{TEAL} & 0&  0.6M& 128.8 & 38.1  \\
\hline
\end{tabular}
\caption{Model performance on the COCO2014 test set. The results of the baselines are cited from their papers directly.}
\label{Caption}
\end{table}

\begin{table*}[t]
\centering
\begin{tabular}{l|ccc|ccc|cc|c}
\hline
          \multicolumn{1}{c}{\multirow{2}{*}{Method}} & \multicolumn{3}{c}{\textbf{Subject}} & \multicolumn{3}{c}{\textbf{Conext Modality}} & \multicolumn{2}{c}{\textbf{Grade}} & \multicolumn{1}{c}{\multirow{2}{*}{Average}}  \\
          \multicolumn{1}{c|}{}  & NAN & SOC & LAN  & TXT & IMG &NO & G1-6 & G7-12 &\multicolumn{1}{c}{}               \\ \hline
          LLaMA-Adapter  & 84.37& 88.30& 84.36& 83.72& 80.32& 86.90& 85.83& 84.05 & 85.19\\ \hdashline
          Human  & 90.23& 84.97& 87.48& 89.60& 87.50& 88.10&91.59 &82.42 & 88.40 \\
          GPT-3.5 & 74.64& 69.74& 76.00& 74.44& 67.28& 77.42& 76.80& 68.89& 73.97 \\
          GPT-3.5 w/ COT &75.44 & 70.87& 78.09& 76.48& 67.43& 79.93& 78.23 & 69.68& 75.17\\
          $\text{MM-COT}_{base}$  & 87.52& 77.17 & 85.82& 87.88& 82.90& 86.83& 84.65& 85.37& 84.91 \\
          $\text{MM-COT}_{large}$  & 95.91 & 82.00& 90.82& 95.26& 88.80 & 92.89& 92.44& 90.31& 91.68\\
          LLaVA-7B  & - & - & - & - &  - & - & - & - & 89.84 \\ 
          LLaVA-13B  & 90.36 & 95.95& 88.00& 89.49& 88.00& 90.66& 90.93& 90.90& 90.92\\ \hdashline
          \textbf{TEAL (Ours)} & 89.00 & 92.94& 86.42&85.06 & 83.00& 88.92& 86.26& 84.90& 87.12\\ \hline

\end{tabular}
\caption{Results on the ScienceQA test set. For the baselines, we directly cite the results from their papers.}
\label{scienceQA_result}
\end{table*}

\subsection{ScienceQA} 
ScienceQA \citep{lu2022learn} is collected from elementary and high school science curricula and contains 21,208 multimodal multiple-choice science questions. Out of the questions in ScienceQA, 10,332 (48.7\%) have an image context, 10,220 (48.2\%) have a text context, and 6,532 (30.8\%) have both. ScienceQA has rich domain diversity across 3 subjects, 26 topics, 127 categories, and 379 skills, and the benchmark dataset is split into training, validation, and test splits with 12726, 4241, and 4241 examples, respectively. The main baseline that can be used to make a fair comparison with our method is the LLaMA-Adapter \citep{zhang2023llama}. We also cite the results of two representation methods (GPT-3.5 and GPT-3.5 w/ COT)  \citep{lu2022learn},  one multi-modal COT method (MM-COT) \citep{zhang2023multimodal}, human evaluation \citep{lu2022learn}, and LLaVA \citep {liu2023visual} which tunes the full parameters of the vicuna with large-scale multi-modal pre-training corpus. Table \ref{scienceQA_result} presents the experimental results. As shown in Table \ref{scienceQA_result}, we can find that the proposed \textit{TEAL} achieves about 2 points improvement on average compared to the baseline of LLaMA-Adapter. 

\begin{wraptable}{r}{0.55\textwidth}\small  
\centering
\begin{tabular}{c|c}
\hline
Model & WER \\ \hline
$\text{HuBERT}_{large}$ \citep{hsu2021hubert} & 31.77 \\ 
$\text{Whisper}_{small}$ \citep{radford2023robust} & 18.8 \\ \hdashline
$\text{Whisper}_{small}$ + LLaMa-Adapter & 26.96 \\  \hdashline
\textbf{TEAL (Ours)} & 24.22 \\ \hline
\end{tabular}
\caption{Results on the CoVoST 2 ASR test set. }
\label{CoVoST_result}
\end{wraptable}

\subsection{CoVoST 2}
For audio, we conduct experiments on the CoVoST 2 \citep{wang2020covost} ASR English dataset, which contains 232976 audio-text training pairs, 15532 validation pairs, and 15532 test pairs. 
We use the word error rate (WER) as the metric. 
We implement the audio tokenizer by applying k-means clustering on the 11th layer of $\text{Whisper}_{small}$.\footnote{We tested different layers of $\text{Whisper}_{small}$ and obtained the best performance on 11th layer.} The number of cluster centers is set as 8192 and the effect of the number of cluster centers will be investigated in Section \ref{cluster_center}. While training and inference, the audio and the corresponding prompt will be processed into token sequences and fed into the MM-LLM directly. For a fair comparison, our main baseline is also implemented based on LLaMa-Adapter and $\text{Whisper}_{small}$, where the $\text{Whisper}_{small}$ is utilized as an encoder to extract the dense audio features from the raw audio waves. We use the default adapter architecture to integrate the audio features into the MM-LLM.
As Table \ref{CoVoST_result} shows, combining an audio tokenizer makes LLM possess better multi-modal understanding ability than explicitly integrating an audio encoder, with a WER score improvement of 2.74. This may be because that having modalities in the same token format makes it easier to integrate multi-modal information for LLM. 
 
\subsection{Image Generation}
\begin{figure}
    \centering
    \resizebox{1.0\textwidth}{!}{
    \includegraphics{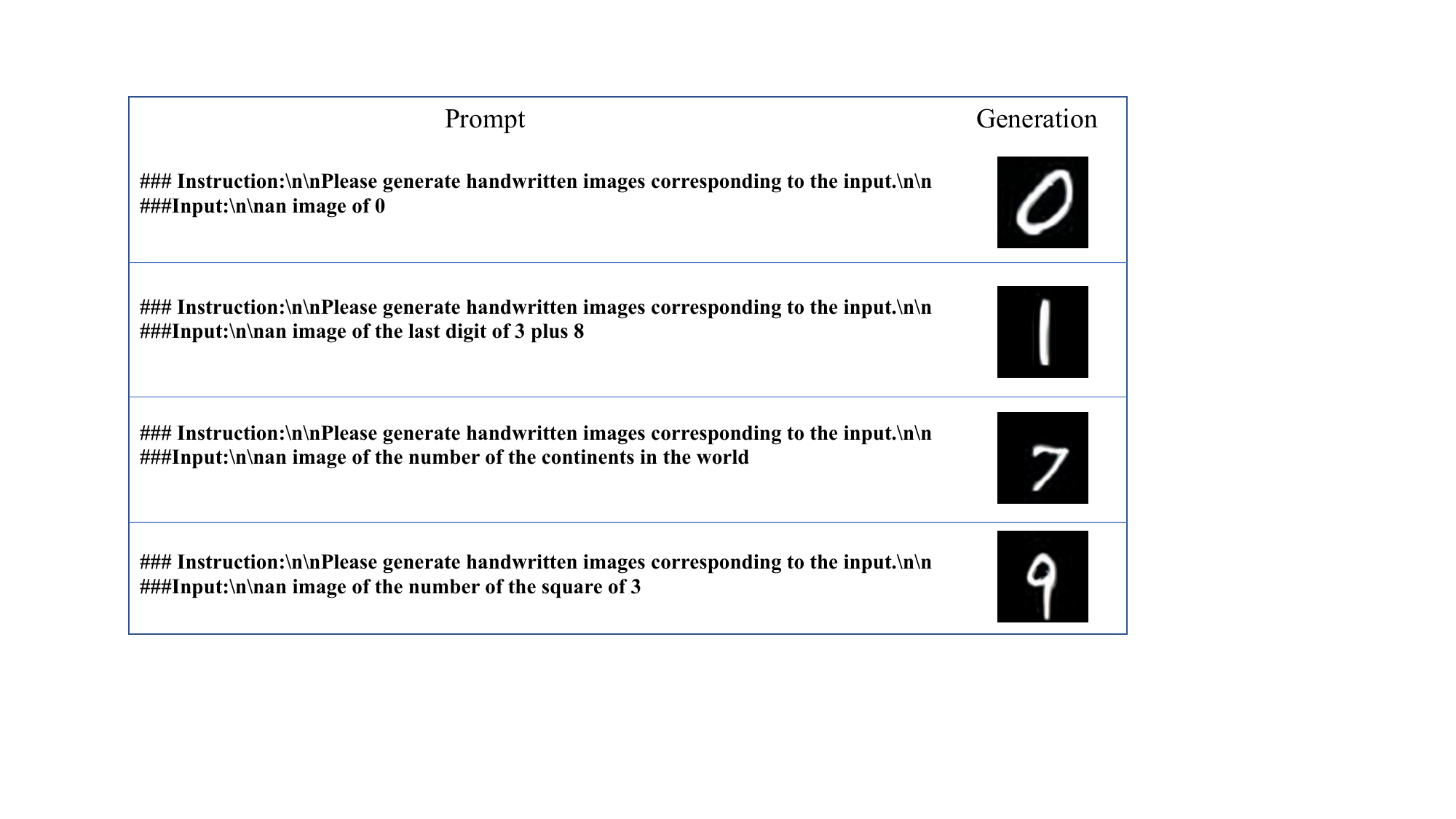}}
    \caption{Some examples of the text-to-image generation on MNIST test set. We test with both simple and complex questions for the proposed \textit{TEAL}.}
    \label{mnist_case}
\end{figure}

Following \citep{yu2023spae}, we show several text-to-image generation examples on the MNIST dataset \citep{deng2012mnist} in Figure \ref{mnist_case}. Different from \citep{yu2023spae}, we do not use any prompt example for in-context learning. As the BEiT-V2 is not good at image reconstruction, we apply the VQGAN as the tokenizer for image generation.\footnote{This is because the BEiT-V2 is not trained to reconstruct the image but to recover the prediction of its teacher model.} From Figure \ref{mnist_case}, we can find that the proposed \textit{TEAL} empowers the frozen textual LLM with the ability to generate the image following the prompt query. We also test with complex questions requiring mathematical reasoning or common sense knowledge, and the model is able to give the right responses. These results show that \textit{TEAL} not only learns how to generate non-textual content but also maintains its previous ability in textual understanding. We notice that the quality of the generated image is not so perfect, and we leave the work of polishing the quality of generated images in the next version.

\section{Analysis and Discussion}
\label{Results}
\begin{wraptable}{r}{0.5\textwidth}\small
\centering
\begin{tabular}{l|cc|c}
\hline
   \multicolumn{1}{c}{\multirow{2}{*}{Model}} & \multicolumn{2}{c}{COCO Caption} & \multicolumn{1}{c}{\multirow{2}{*}{ScienceQA (ave.)}}\\
   \multicolumn{1}{c|}{} & CiDER&	BLEU-4  &  \multicolumn{1}{c}{}    \\
    \hline
 DALLE & 110.8 & 23.9 & 77.12 \\
VQGAN & 117.5 & 26.1 & 79.56\\
\textbf{BEiT-V2}  & \textbf{130.1} & \textbf{37.6} & \textbf{88.00} \\
\hline
\end{tabular}
\caption{The performance of different tokenizers on the validation sets of the COCO2014 and ScienceQA. We keep all parameters and data the same and only vary the tokenizers.}
\label{image_tokenize}
\end{wraptable}

\subsection{Different tokenizers}
We show how the tokenizer affects the performance by testing different tokenizers for the image and audio. For the image, we report the performance on the validation set of COCO-caption by varying the image tokenizers. Results are shown in Table \ref{image_tokenize}. We find that different tokenizers result in significant differences in the final performance, and BEiT-V2 achieves the best result. Compared to the baseline of VQ-GAN, BEiT-v2 achieves 11.5 BLEU points improvement on the task of COCO-caption and 8.5 accuracy points on ScienceQA. The significant performance gap highlights the importance of the tokenizer. We speculate that the main reason for BEiT-v2 achieving such a significant advantage is that BEiT-v2 has acquired much semantic information during its pre-training, and the semantic information in the tokenizer is crucial for aligning different modalities. 

We have similar observations in the modality of audio. We have tried different tokenizers such as HuBERT Clustering, $\text{Whisper}_{small}$ Clustering. 
Table \ref{audio_tokenize} shows the comparison. We also list some CoVoST2 ASR results with different tokenizers of AudioPaLM \citep{rubenstein2023audiopalm} to make a comparison. Both the experiments of AudioPaLM and TEAL demonstrate that different tokenizers can have a significant impact on performance. A good tokenizer is crucial, and it is an area worth exploring for future work.

\begin{table}
\centering
\begin{tabular}{l|c|c|c|c}
\hline
Tokenizer & Type & LLM & LLM size& WER \\ \hline
W2V-BERT\citep{chung2021w2v} & Cluster & PaLM & 8B & 50.1 \\ 
USM-v1\citep{zhang2023google} & Cluster & PaLM & 8B & 40.2 \\ 
USM-v2\citep{zhang2023google} & Cluster & PaLM & 8B & 22.3 \\ \hline

HuBERT\citep{hsu2021hubert} & Cluster & LLaMa & 7B & 56.2 \\ 
$\text{Whisper}_{small}$ \citep{radford2023robust}& Cluster & LLaMa & 7B & 24.2 \\  \hline
\end{tabular}
\caption{The performance of different tokenizers on the test sets of the CoVoST 2. }
\label{audio_tokenize}
\end{table}

\begin{wraptable}{r}{0.5\textwidth}\small
\centering
\begin{tabular}{l|c|c|c|c}
\hline
Vocab Size & 1024 & 2048 & 4096 & 8192  \\ \hline
WER & 40.22 & 30.85 & 25.31 & \textbf{21.49} \\ \hline
\end{tabular}
\caption{We randomly sample 500 audio-text pairs from the development set of the CoVoST 2, and the performance with different vocab sizes is shown in the table. }
\label{audio_vocab_size}
\end{wraptable}


\subsection{K-means Cluster analysis}
\label{cluster_center}
Table \ref{audio_vocab_size} shows the difference when adopting different audio vocab sizes. All the tokenizers are trained based on the features of the 11th layer of $Whisper_{small}$. We find out that the vocab size has a substantial effect on performance. Compared to clustering 1024 tokens, clustering 8192 tokens can result in a WER improvement of over 18 percentage points. This makes the clustering-based discretization approaches more versatile than the VQ-based neural codecs for the audio. The former can adjust the vocabulary size by tuning the number of clustering centers, while the latter needs to retrain a vector quantization module. 

\begin{wraptable}{r}{0.7\textwidth}\small
\centering
\begin{tabular}{l|cc|c}
\hline
   \multicolumn{1}{c}{\multirow{2}{*}{Model}} & \multicolumn{2}{c}{COCO Caption} & \multicolumn{1}{c}{\multirow{2}{*}{ScienceQA (ave.)}}\\
   \multicolumn{1}{c|}{} & CiDER&	BLEU-4  &  \multicolumn{1}{c}{}    \\
    \hline
 \textbf{TEAL (Ours)}  & \textbf{130.1} & \textbf{37.6} & \textbf{88.00} \\
w/o 1st-stage finetuning & 127.8& 35.4 & 86.19\\
w/o embedding initialization  &129.1  & 36.2 &86.82  \\
w/o bias-norm tuning & 126.9 & 35.7 & 85.74 \\
\hline
\end{tabular}
\caption{Ablation study on the proposed model. `w/o 1st-stage finetuning' indicates that the model is trained with the 2nd-stage finetuning directly. `w/o embedding initialization' means that we initialize the word embedding and output matrix randomly. `w/o bias-tuning' means that the parameters of bias and norm are not added during the 2nd stage finetuning.}
\label{ablation_study}
\end{wraptable}

\subsection{Ablation study}
To investigate the significance of each module in our model and method, we conduct an ablation study by training multiple versions of our model with some missing components, i.e., the 1st-stage finetuning, the embedding initialization, and the bias-norm tuning. We report the performance on the validation sets and Table \ref{ablation_study} lists the experimental results. From Table \ref{ablation_study}, we can find that the best performance is obtained with the simultaneous use of all the tested components. The most critical components are the bias-norm tuning and the 1st-stage finetuning, which shows that the training strategies need to be carefully devised to ensure high performance. A surprising phenomenon is that when we randomly initialize the word embedding (`w/o embedding initialization' in Table \ref{ablation_study}), we do not observe a significant performance decrease. This result suggests that it is the way the tokenizer discretizes the image, rather than the word embedding preserved in the tokenizer, critical to the final performance. The reason why random initialization causes a certain degree of performance decrease is likely due to the relatively small size of the training data. We speculate that when the amount of training data reaches a certain level, the performance gap will disappear.

\section{Conclusion and Future work}
In this paper, we propose \textit{TEAL}, an approach to training a fully token-in-token-out MM-LLM by treating the input from any modality as a token sequence and learning a joint embedding space for all modalities. \textit{TEAL} empowers the frozen textual LLM with the ability to perform understanding and generation involving non-textual modalities. Extensive experiments show that, compared to the baseline models which integrate non-textual encoders, our approach achieves superior performance on non-textual understanding tasks, and paves a simple way for non-textual generation.

There are two main promising directions for the future work. Firstly, we are interested in constructing an MM-LLM model that can handle more tasks and more modalities. The token-in-token-out architecture has the potential to handle all tasks in AI within one model. Secondly, we want to devise a general tokenizer, which can discretize the input from textual and non-textual modalities in a unified way. With such a general tokenizer, aligning the samples from different modalities is simpler and more straightforward.

\clearpage

\bibliography{iclr2024_conference}
\bibliographystyle{iclr2024_conference}

\appendix
\clearpage
\section{Prompts for different tasks}
\label{sec_prompt}
We present the prompts we use for different tasks in Table \ref{prompts}, which are generated by GPT4 automatically.

\begin{table}
\centering
\begin{tabular}{c|c}
\hline
Task & Prompts \\ \hline
 \multicolumn{1}{c|}{\multirow{10}{*}{image caption}} & Please provide a caption for the image that has been given. \\ 
  \multicolumn{1}{c|}{\multirow{10}{*}{}} & Your task is to write a caption for the provided image. \\
   \multicolumn{1}{c|}{\multirow{10}{*}{}} & The objective is to come up with a caption for the image that has been provided. \\
    \multicolumn{1}{c|}{\multirow{10}{*}{}} & You are required to write a caption for the provided image. \\
     \multicolumn{1}{c|}{\multirow{10}{*}{}} &  Your job is to create a caption for the image that has been given. \\ 
      \multicolumn{1}{c|}{\multirow{10}{*}{}} & The challenge is to think of a caption for the provided image. \\
        \multicolumn{1}{c|}{\multirow{10}{*}{}} & You have been given an image and your goal is to write a caption for it. \\
          \multicolumn{1}{c|}{\multirow{10}{*}{}} & You have been given an image and your task is to write a caption for it. \\
            \multicolumn{1}{c|}{\multirow{10}{*}{}} &The task at hand is to provide a caption for the image that has been provided. \\
              \multicolumn{1}{c|}{\multirow{10}{*}{}} & Your assignment is to come up with a caption for the provided image. \\ \hline
   ASR & Write a response that appropriately completes the request based on the provided audio.\\ \hline       
 \multicolumn{1}{c|}{\multirow{10}{*}{image generation}} & Create an image that perfectly matches the input sentence. \\ 
  \multicolumn{1}{c|}{\multirow{10}{*}{}} &  Generate an image that fits the input sentence perfectly. \\ 
   \multicolumn{1}{c|}{\multirow{10}{*}{}} & Produce an image that seamlessly complements the input sentence. \\ 
    \multicolumn{1}{c|}{\multirow{10}{*}{}} & Create a picture that perfectly corresponds to the input sentence. \\ 
     \multicolumn{1}{c|}{\multirow{10}{*}{}} & Generate an image that perfectly aligns with the input sentence. \\ 
      \multicolumn{1}{c|}{\multirow{10}{*}{}} & Create an image that perfectly harmonizes with the input sentence. \\ 
       \multicolumn{1}{c|}{\multirow{10}{*}{}} & Produce an image that perfectly integrates with the input sentence. \\ 
        \multicolumn{1}{c|}{\multirow{10}{*}{}} & Generate an image that perfectly suits the input sentence. \\ 
         \multicolumn{1}{c|}{\multirow{10}{*}{}} & Create an image that perfectly matches the input sentence in every way. \\ 
          \multicolumn{1}{c|}{\multirow{10}{*}{}} & Produce an image that perfectly corresponds to the input sentence in every aspect. \\ \hline

\end{tabular}
\caption{The prompts generated by GPT4 for different tasks.}
\label{prompts}
\end{table}

\end{document}